%% file: main.tex
\documentclass[10pt,onecolumn,letterpaper]{article}
\usepackage{cvpr}
\usepackage[numbers,sort&compress]{natbib}
\usepackage{times}
\usepackage{eso-pic}

\usepackage{epsfig}
\usepackage{graphicx}
\usepackage{amsthm}
\usepackage{amsmath}
\usepackage{amssymb}
\usepackage[breaklinks=true,bookmarks=false]{hyperref}

\numberwithin{equation}{section}
\usepackage{datetime}

\usepackage[utf8]{inputenc} % allow utf-8 input
\usepackage[T1]{fontenc}    % use 8-bit T1 fonts
\usepackage{hyperref}       % hyperlinks
\usepackage{url}            % simple URL typesetting
\usepackage{booktabs}       % professional-quality tables
\usepackage{amsfonts}       % blackboard math symbols
\usepackage{nicefrac}       % compact symbols for 1/2, etc.
\usepackage{microtype}      % microtypography
\usepackage{xcolor}         % colors

\usepackage{graphicx}
\usepackage{amsmath} 
\usepackage{algorithm}
\usepackage{algorithmic}
\usepackage{colortbl} 
\usepackage{tcolorbox}

\usepackage{multirow}
\usepackage{array}
\usepackage{longtable}
\usepackage{adjustbox}
\usepackage{array}
\usepackage{lscape}
\usepackage{natbib}
\usepackage[a4paper,margin=1in,headheight=15pt,headsep=24pt]{geometry}

\usepackage{fancyhdr}

\input{math_commands.tex}

\usepackage{hyperref}
\usepackage{url}
\usepackage{graphicx}
\usepackage{multirow}
\usepackage{booktabs}
\usepackage{array}
\usepackage[table]{xcolor}
\usepackage{pifont}
\usepackage{subfigure}

\usepackage{algorithm}
\usepackage{algorithmic}
\usepackage{amsmath,bm}
\usepackage{amssymb}
\usepackage{amsthm}
\usepackage{multirow} 
\usepackage{graphicx}
\usepackage{amsmath}
\usepackage{amsthm}

\usepackage{enumitem}

\cvprfinalcopy % *** Uncomment this line for the final submission
\def\cvprPaperID{****} % *** Enter the CVPR Paper ID here
\allowdisplaybreaks

\begin{document}
\lhead{}
\lfoot{\date{\today},\date{\currenttime}}
\rfoot{NGD for DL}

% \title{Instruction Learning Paradigms: A Dual Perspective on White-box and Black-box LLMs}
\title{SPOGW: a Score-based Preference Optimization method via Group-Wise comparison for workflows}
\author{
% Yanwei Ren, Liu Liu, Baosheng Yu, Jiayan Qiu, Quan Chen
\textbf{Yitong Cui}$^{1}$\quad
\textbf{Liu Liu}$^{2,1*}$\quad
\textbf{Baosheng Yu}$^{3}$\quad
\textbf{Jiayan Qiu}$^{4}$\quad
\textbf{Xikai Zhang}$^{1}$\quad\\
\textbf{Likang Xiao}$^{1}$\quad
\textbf{Yixing Liu}$^{5}$\quad
\textbf{Quan Chen}$^{6}$\\[0.5em]
$^1$Hangzhou International Innovation Institute and , Beihang University\\
$^2$School of Artificial Intelligence, Beihang University\\
$^3$ Nanyang Technological University\quad
$^4$University of Leicester \\
$^4$China Mobile Communications Company Limited Research Institute\\
$^5$Kuaishou Technology
}
\maketitle
\begingroup
\renewcommand\thefootnote{*}
\footnotetext{Corresponding author: \texttt{liuliubh@buaa.edu.cn}}
\endgroup
\input{MainText/0abs}
\input{MainText/1Intr}

\input{MainText/2RelatedWork}

\input{MainText/3Met}

\input{MainText/4Exp}

\input{MainText/6Con}

% \section*{References}

\medskip

% \today
{\small
 \bibliographystyle{ieee}
 \bibliography{main}
}

%%%%%%%%%%%%%%%%%%%%%%%%%%%%%%%%%%%%%%%%%%%%%%%%%%%%%%%%%%%%
% \input{NList}
\end{document}

%% file: math_commands.tex
\usepackage{amsmath,amsfonts,bm}

\def\eqref#1{equation~\ref{#1}}
\def\1{\bm{1}}

\DeclareMathAlphabet{\mathsfit}{\encodingdefault}{\sfdefault}{m}{sl}
\SetMathAlphabet{\mathsfit}{bold}{\encodingdefault}{\sfdefault}{bx}{n}

%% file: MainText/0abs.tex
\begin{abstract}
\quad \quad Large language models (LLMs) have exhibited significant capabilities in addressing challenging problems throughout various fields, often through the use of agentic workflows that adhere to structured instructions and multi-step procedures. However, designing such workflows demands substantial manual effort, posing challenges to scalability and generalizability. Recent studies have aimed to minimize the human intervention needed for their construction, leading to advances in automated techniques for optimizing agentic workflows. However, current approaches are often constrained by their limited representational capacity, insufficient adaptability, weak scalability, and pairwise comparison paradigm—issues that stem primarily from a dependence on discrete optimization techniques. To overcome these limitations, we introduce a new score-based preference approach, refereed as SPOGW, which operates directly on cardinal reward signals through group-wise comparison and enables more efficient and stable optimization in a continuous space. SPOGW incorporates Iterative offline GRPO (ioGRPO) with advantage-masked KL divergence (mKL), which regulates training update by placing greater emphasis on the advantageous regions of the policy response. In five benchmark datasets covering mathematical reasoning, coding, and question answering, SPOGW matches or exceeds the performance of current state-of-the-art approaches, presenting a viable and forward-looking methodology for automated generation and optimization of agentic workflows.
\end{abstract}

%% file: MainText/1Intr.tex
\section{Introduction}
Large Language Models (LLMs) have demonstrated versatile capabilities in addressing challenging tasks across numerous domains, such as data interpretation, code generation, mathematical problem solving, and question answering~\citep{liu2024survey},~\citep{li2024dawn},~\citep{zhong2024debug},~\citep{wang2024chain},~\citep{xu2023lemur}. However, the progress of LLM-based systems is considerably dependent on hand-crafted agentic workflows—predefined sequences of LLM calls coupled with precise instructions. The substantial human effort involved in developing and refining these workflows impedes scalability, restricts adaptability to novel or intricate scenarios, and complicates knowledge transfer between different tasks~\citep{tang2023verifai}.
A key research direction that has thus gained traction aims to overcome the constraints of static workflows through automated techniques for systematically generating and refining workflows. Such optimizations may be applied at multiple levels, such as improving prompts, adjusting hyperparameters, or redesigning the workflow architecture itself~\citep{chen2023autoagents},~\citep{hu2024automated},~\citep{song2024adaptive},~\citep{zhang2024aflow},~\citep{li2024autoflow},~\citep{zhang2024g}.

Current automated optimization techniques are often limited by predefined structural templates and inflexible representations of the workflow space~\citep{khattab2024dspy},~\citep{liu2024dynamic},~\citep{yuksekgonul2024textgrad},~\citep{zhuge2023mindstorms}. For instance, while DyLAN~\citep{liu2024dynamic} deliberately designs the communication protocol for LLM-based debates, it does not explore alternative interaction patterns. GPTSwarm~\citep{zhuge2023mindstorms} utilizes graph representations and applies reinforcement fine-tuning for improvement, yet its failure to account for conditional states within graphs inherently constrains the explorable solution space.

To enhance the expressiveness and adaptability of workflow representations, approaches like ADAS~\citep{hu2024automated}, Aflow~\citep{zhang2024aflow}, and ScoreFlow~\citep{wang2025scoreflow} utilize workflow representations that are based on code. However, ADAS is hampered by the accumulation of irrelevant data and the increasing complexity during optimization, ultimately reducing its performance. Aflow improves the representation of workflows via code by incorporating a core element known as the named node, which encapsulates settings for LLM invocations to enable detailed modeling. The method also includes dedicated operators that carry out predefined logic for composing nodes. However, the effectiveness of Aflow's optimization based on Monte Carlo Tree Search is constrained by early convergence, and its discrete nature hinders scalability. ScoreFlow incorporates the Direct Preference Optimization (DPO)~\citep{rafailov2023direct} RL technique into workflow optimization and adapts it to incorporate quantitative feedback. However, its optimization framework is severely constrained by a strong dependence on pairwise preference data, leading to rigidity. Essentially, it requires reframing performance assessment as a binary comparison process instead of directly optimizing a continuous metric of performance, thus hindering its ability to inherently integrate cardinal reward signals.

% To address these challenges, we propose \textbf{SPGW}, a \textbf{S}core-based \textbf{P}reference optimization method via \textbf{G}roup-\textbf{W}ise comparison. SPGW operates directly on cardinal reward signals and performs optimization in a continuous space, overcoming the limitations of pairwise preference paradigms. Our approach consists of two key components:
To address these challenges, we introduce SPOGW, a Score-based Preference Optimization method via Group-Wise comparison. SPOGW directly leverages cardinal reward signals and conducts optimization in a continuous space, thereby overcoming the inherent limitations of traditional pairwise preference paradigms. Our approach is built upon two key components:
\begin{itemize}[leftmargin=0.7cm]
    \item \textbf{Iterative Offline GRPO (ioGRPO)}, which decouples data collection from policy updates. By performing offline sampling and reward acquisition before training, ioGRPO eliminates the instability caused by on-the-fly code execution and API calls during optimization. The process runs in iterative cycles, where each iteration uses the previous checkpoint as the both old and reference policy for generating new training data.
    \item \textbf{Advantage-Masked KL Restriction (mKL)}, which selectively applies KL divergence penalties only to advantageous responses (those with positive advantage values). This ensures the policy stays aligned with high-quality behaviors from the reference model while avoiding unnecessary constraints from low-quality outputs.
\end{itemize}

\begin{figure}[t]
\begin{center}
\includegraphics[width=0.9\textwidth]{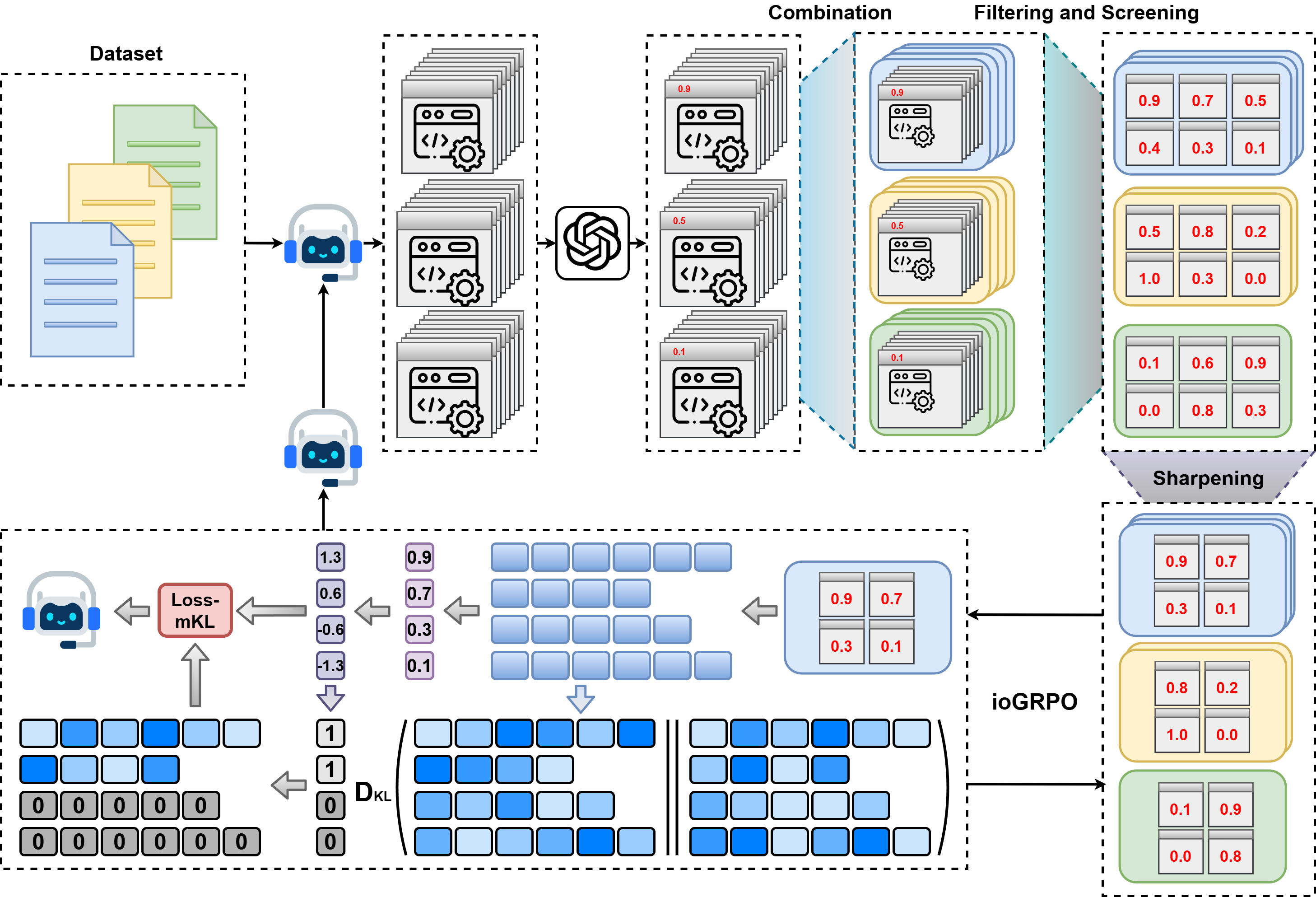}
 % \vskip -0.25in
%\framebox[4.0in]{$\;$}
% \fbox{\rule[-.5cm]{0cm}{4cm} \rule[-.5cm]{4cm}{0cm}}
\end{center}
\caption{\textbf{Pipeline of SPOGW:} The framework generates multiple workflows for each query, then executes and evaluates each workflow to obtain a score, and then conducts combination and subsequent group-wise data processing,  which feeds into ioGRPO optimization cycle.
% ,  performing policy updates with advantage-masked KL restriction, producing an improved policy for the next iteration.
}
% \vskip -0.25in
\end{figure}

Together, these innovations enable more stable, efficient, and scalable optimization of agentic workflows. Experiments across mathematical reasoning, coding, and QA benchmarks show that SPOGW matches or surpasses state-of-the-art methods, demonstrating its effectiveness as a general-purpose framework for automated workflow generation and optimization.

%% file: MainText/2RelatedWork.tex
\section{Related works}
\subsection{Reinforcement Learning for Advanced Reasoning in LLMs}
The integration of reinforcement learning (RL) to improve LLM reasoning has attracted considerable attention~\citep{cheng2025revisiting},~\citep{zhang2025critique},~\citep{xiong2025minimalist}, owing to its ability to enable self-improvement without dependence on human-annotated solutions. This is commonly realized through fine-tuning on sophisticated reasoning problems, with the objective of promoting varied reasoning strategies~\citep{gandhi2025cognitive},~\citep{yue2025does}. Notable advances such as OpenAI o1~\citep{jaech2024openai} and DeepSeek-R1~\citep{guo2025deepseek} illustrate that RL methods can be effectively applied to large-scale commercial systems, substantially pushing the boundaries of reasoning performance and unveiling emergent skills including extended reasoning chains.
Recent progress has employed reinforcement learning guided by scalar feedback signals~\citep{jaech2024openai},~\citep{guo2025deepseek},~\citep{liu2025understanding},~\citep{yu2025dapo}. For instance, a positive reward (e.g., +1) may be assigned for accurate outputs, while a negative penalty (e.g., –1) is applied to erroneous responses to provide clear learning signals.
Common algorithmic choices in this line of work include online policy optimization techniques such as REINFORCE~\citep{10.1007/BF00992696}, Proximal Policy Optimization (PPO)~\citep{schulman2017proximal}, Group Relative Policy Optimization (GRPO)~\citep{shao2024deepseekmath}, and Decoupled Clip and Dynamic Sampling Policy Optimization (DAPO)~\citep{yu2025dapo}.
Although powerful within their respective specialized domains, current research on applying RL algorithms to workflow optimization techniques remains relatively scarce. Furthermore, the adaptability and effectiveness of existing RL algorithms for workflow optimization tasks lack sufficient empirical validation and analysis.

\subsection{Automated Agentic Workflow Optimization}
\paragraph{Agentic Workflows}
Agentic workflows and autonomous agents~\citep{zhuge2023mindstorms},~\citep{hong2024data},~\citep{wang2023voyager} constitute two principal paradigms for applying LLMs. The former operate through fixed, predefined sequences of steps—orchestrating multiple calls to LLMs—to accomplish tasks in a structured manner. In contrast, the latter tackle problems adaptively via self-directed reasoning and action. Unlike autonomous agents, which often demand carefully crafted action spaces and decision rules tailored to particular environments, agentic workflows can be built upon accumulated human expertise and refined through iteration. This positions them as a more automatable and scalable approach for many practical applications.
\paragraph{Automated Workflow Optimization}
Workflow optimization techniques~\citep{zhou2024symbolic},~\citep{hu2024automated},~\citep{zhang2024aflow},~\citep{li2024autoflow},~\citep{zhang2024g},~\citep{wang2025scoreflow} aim to improve the structural design of workflows, enhancing their robustness across varied tasks. However, their effectiveness is often constrained by rigid representations—for instance, the loss of conditional logic in graph-based structures—which narrows the search space and limits adaptability to complex scenarios. To overcome these limitations, methods such as ADAS~\citep{hu2024automated}, Aflow~\citep{zhang2024aflow} and  ScoreFlow~\citep{wang2025scoreflow} employ code-based workflow representations.
% However, ADAS suffers from the accumulation of extraneous information and growing complexity in the optimization process, which ultimately impair its operational effectiveness.
Aflow enhances code-based workflow representation by introducing a foundational component called named node, which packages parameters for LLM calls to support fine-grained workflow modeling. 
% The approach also integrates specialized operators that execute predefined compositional logic between nodes.
However, the efficacy of Aflow's Monte Carlo Tree Search-based optimization is limited by premature convergence, while its discrete optimization property impedes scalability.
ScoreFlow integrates the Direct Preference Optimization (DPO)~\citep{rafailov2023direct} reinforcement learning method into workflow optimization and extends it to account for quantitative feedback. 
However, its optimization paradigm is critically limited by its heavy reliance on pairwise preference data, resulting in inflexibility. 
% Inherently, it requires recasting performance evaluation as a discrete, pairwise comparison task rather than directly optimizing a continuous, absolute performance metric, which prevents it from natively incorporating or utilizing cardinal reward signals.

%% file: MainText/3Met.tex
\section{Methods}
\subsection{Score-based Preference Dataset}

\paragraph{Data construction}
% \textbf{Data construction}
In the score-based workflow application, 
for each query $q$ in the dataset $\mathcal{D}$, a generator LLM produces $m$ corresponding workflows (denoted as $g_i(q)$, $i\in[m]$). Executing each workflow $g_i(q)$ yields a result for each query $q$. Subsequently, these results are evaluated to produce the corresponding scores $s_i$ ($s_i \in [0, 1]$). In experiments, the workflows are executed by independently querying an executor LLM;
% the resulting outputs are evaluated using metrics such as the average F1 score or the win rate to determine $s_i$. 
Unlike self-improvement methods~\citep{jiang2024self} that employ the generator model for evaluation, score-based workflow~\citep{wang2025scoreflow} leverages external resources (e.g., a validation dataset and an executor LLM) to realize the self-referential property of the iterative process.
However, ScoreFlow~\citep{wang2025scoreflow} is limited by the pairwise comparison paradigm, which constructs score-based workflows into preference pairs, thereby restricting the scale of comparison samples and lacking flexibility and scalability.

Thus, we construct the group-wise training data set for query $q$ from a new perspective, with a initial group size of $n$ ($n \leq m$) pairs. For each data instance, we define
\begin{align}
    D_q = \left\{ q,  (g_1(q),s_1), \ldots, (g_n(q),s_n) \} \right\},
\end{align}
where consists of a query $q$, the corresponding workflows $g_j(q)$, and their respective scores $s_j$, $j\in[n]$. Since $m$ workflows are initially generated for each query, a total of $M$ group-wise training instances can be created for each query $q$ through combinations, i.e., $M=C(m, n)$, which represents the number of ways to choose n distinct elements from a set of m elements. Thus, we obtain a group training set for query $q$, i.e., $\mathcal{D}_q = \{ D_q^1, \ldots, D_q^{M} \}$. Finally, the complete preprocessed group-wise training dataset $\mathcal{D}_{\text{pre}}$ is formed by aggregating the data from all queries, i.e., $\mathcal{D}_{\text{pre}} = \bigcup_{q \in \mathcal{D}} D_q$.

\paragraph{Filtering and Screening}
The quality of the initial group-wise dataset $\mathcal{D}_{\text{pre}}$ is heterogeneous, with significant variation across samples. Certain instances suffer from highly similar sampled responses and nearly identical intra-group reward scores. Utilizing these suboptimal samples would detrimentally impact the efficacy of advantage estimation. Therefore, we design a subsequent data curation pipeline to post-process the raw dataset. The objective of this pipeline is to yield a refined dataset characterized by superior \textit{intra-group diversity} and clearer \textit{distinction between high- and low-quality responses}. This high-quality data enables the advantage calculation to produce a stronger and more unambiguous learning signal, thereby enhancing the efficiency of the reinforcement learning process and ultimately leading to improved final performance.

% Finally, the constructed group-wise training dataset undergoes a sequential process of filtering, screening, and group sharpening to enhance its quality and effectiveness for policy optimization.

Specifically, the screening process is applied to this filtered set. For each remaining instance, the variance of the reward scores $\text{Var}({s_1, ..., s_n})$ for its $n$ responses is computed. The instances are then ranked in descending order based on this variance. We select the top-$N$ instances with the highest variance for inclusion in the final training set (or all instances if the total number is less than $N$). A high variance indicates a reward distribution with sufficient distinction among responses, enabling the model to more effectively learn nuanced preferences. This screening procedure ensures \emph{intra-group diversity} and mitigates the risk of advantage calculation failure due to reward homogenization.

\paragraph{Group Sharpening} 
% In order to XXXX, we conduct the curateed step, which applied to the screened dataset, called \emph{group sharpening}. 
To achieve a higher effective variance and a more polarized reward distribution using fewer samples, we perform a curated step on the screened dataset, called \emph{group sharpening}.
For a given data instance $D_q$, we first sort the responses and their corresponding rewards in ascending order based on the reward value, resulting in the same data instance but under different orders, i.e.,
$\hat{D}_q = \left\{ q,  (g_{(1)}(q),s_{(1)}), \ldots, (g_{(n)}(q),s_{(n)}) \} \right\}$,
% ${q, {y_{(1)}, ..., y_{(n)}}, {R_{(1)}, ..., R_{(n)}}}$ 
where $s_{(1)} \leq ... \leq s_{(n)}$.
The sharpening operation then retains only the top-$t$ and bottom-$t$ responses, effectively constructing a new, sharper instance:
\begin{align}
    \tilde{D}_q = \left\{ q,  (g_{(1)}(q),s_{(1)}), \ldots,  (g_{(t)}(q),s_{(t)}), (g_{(n-t+1)}(q),s_{(n-t+1)}),\ldots,(g_{(n)}(q),s_{(n)}) \} \right\},
    \label{Equation_1}
\end{align}
%
% \begin{equation}
% {q, {y_{(1)}, ..., y_{(t)}, y_{(n-t+1)}, ..., y_{(n)}}, {R_{(1)}, ..., R_{(t)}, R_{(n-t+1)}, ..., R_{(n)}}},
% \end{equation}
%
where $2t < n$. The group size is thus reduced from $n$ to $2t$.
%
% The purpose of group sharpening is to achieve a higher effective variance and a more polarized reward distribution using fewer samples.
%
%
By focusing on the most positively and negatively rewarded examples, this technique amplifies the contrast within the data, yielding a stronger and clearer learning signal for the advantage estimator. This leads to more stable and efficient policy updates during training.
% Finally, the complete constructed GRPO training dataset undergoes sequential filtering and selection. The filtering criterion removes any training instance where all workflows for a given question have identical scores; otherwise, the instance is retained. This process yields a filtered training dataset. Subsequently, a selection step is applied: instances are ranked in descending order based on the variance of the scores of the workflows for each question. The top-$N$ instances with the highest variance are selected to form the final GRPO training dataset.

\subsection{From GRPO to Iterative offline GRPO}
Group Relative Policy Optimization (GRPO) is an online reinforcement learning algorithm commonly employed for fine-tuning Large Language Models (LLMs). It extends the framework of Proximal Policy Optimization (PPO)~\citep{schulman2017proximal}, while avoiding the requirement for explicit value function estimation by computing advantages through the comparative performance of grouped actions. In the setting of LLM policy optimization, consider a model policy with parameters $\theta$. For every query $q$ from a set $\mathcal{Q}$, a set of candidate responses $\{y_i\}_{i=1}^{n}$ is generated under the previous policy $\pi_{\text{old}}$. These samples are then evaluated by a reward model, producing a corresponding set of rewards $\{R_i\}_{i=1}^{n}$. The objective function for GRPO is expressed as:
% GRPO (Group Relative Policy Optimization) is an online reinforcement learning algorithm widely used for fine-tuning Large Language Models (LLMs). The algorithm builds upon Proximal Policy Optimization (PPO)~\citepp{schulman2017proximal} but eliminates the need for value function approximation by estimating advantages based on relative performance within groups. In the context of LLM policy optimization, assume the model policy is parameterized by $\theta$. For each question $q$ in a given set $Q$, a group of $n$ responses $\{y_i\}_{i=1}^{n}$ is sampled from the old policy $\pi_{\text{old}}$. These responses are then scored by a reward model, producing reward values $\{R_i\}_{i=1}^{n}$. The GRPO training objective is formulated as:
% \begin{align}
%     &\mathcal{J}_{\text{GRPO}}(\theta) \nonumber\\
%     =& \mathbb{E}_{q \sim \mathcal{Q}, \{y_i\}_{i=1}^{n} \sim \pi_{\text{old}}(\cdot | q)} 
%          \frac{1}{n} \sum_{i=1}^{n} \frac{1}{|y_i|} \sum_{t=1}^{|y_i|} \left\{ \min \left[ r_{i,t}(\theta) \hat{A}_{i,t}, \text{clip}(r_{i,t}(\theta), 1 - \epsilon, 1 + \epsilon) \hat{A}_{i,t} \right]\right\}, 
% \end{align}
\begin{align}
    \mathcal{J}_{\text{GRPO}}(\theta)
    = 
         \frac{1}{n} \sum_{i=1}^{n} \frac{1}{|y_i|} \sum_{t=1}^{|y_i|} \left\{ \min \left[ r_{i,t}(\theta) \hat{A}_{i,t}, \text{clip}(r_{i,t}(\theta), 1 - \epsilon, 1 + \epsilon) \hat{A}_{i,t} \right]\right\}, 
\end{align}
% \begin{equation}
% \label{eq:grpo_objective}
% \mathcal{L}^{\text{GRPO}}(\theta) = \mathbb{E} \left[ \frac{1}{n} \sum_{i=1}^{n} \left( \min\left( r_t^{(i)}(\theta) \hat{A}^{(i)}, \text{clip}(r_t^{(i)}(\theta), 1-\epsilon, 1+\epsilon) \hat{A}^{(i)} \right) - \beta \, D_{\text{KL}}\left( \pi_{\theta} \| \pi_{\text{ref}} \right) \right) \right]
% \end{equation}
where the probability ratio $r_{i,t}(\theta)$ is defined as the relative probability of generating a response under the current policy $\pi_{\theta}$ compared to the old policy $\pi_{\text{old}}$ under which the responses were initially sampled:
% where $r_{i,t}(\theta)$ is the probability ratio, comparing the likelihood of the response under the current policy $\pi_{\theta}$ versus the old policy $\pi_{\text{old}}$:
$r_{i,t}(\theta) = \frac{\pi_\theta(y_{i,t} | q, y_{i,<t})}{\pi_{\text{old}}(y_{i,t} | q, y_{i,<t})}$.
% \begin{equation}
% r_{i,t}(\theta) = \frac{\pi_\theta(y_{i,t} | q, y_{i,<t})}{\pi_{\text{old}}(y_{i,t} | q, y_{i,<t})} 
% %, \quad \text{where } r_t^{(i)}(\theta_{\text{old}}) = 1
% \end{equation}
% \begin{equation}
% \label{eq:probability_ratio}
% r_t^{(i)}(\theta) = \frac{\pi_{\theta}(y^{(i)} \mid q)}{\pi_{\text{old}}(y^{(i)} \mid q)}
% \end{equation}
Here, $\epsilon$ represent hyperparameters. 
% The clipping parameter $\epsilon$ restricts the allowable range of the probability ratio, serving to clip overly optimistic estimates and thereby imposing a conservative lower bound to avoid excessively large policy updates. 
% Here, $\epsilon$ and $\beta$ are hyperparameters. The parameter $\epsilon$ controls the range for clipping the probability ratio, enforcing a pessimistic lower bound on policy performance to prevent excessively large policy updates. Simultaneously, $\beta$ modulates the KL divergence penalty, constraining the trained policy from deviating significantly from the reference policy.
The advantage value $\hat{A}_{i,t}$ is computed for all tokens within a response by standardizing the rewards $\{R_i\}_{i=1}^{n}$—specifically, by subtracting the group mean and dividing by the group standard deviation.
% The advantage $\hat{A}_{i,t}$ for all tokens in a response is computed by normalizing the rewards $\{R_i\}_{i=1}^{n}$ using the group mean and standard deviation:
% \begin{equation}
% \label{eq:advantage_calculation}
% \hat{A}_{i,t} = \frac{R_i - \mu}{\sigma}
% \end{equation}
$\hat{A}_{i,t} = \frac{R_i - \mu}{\sigma}$
where $\mu$ and $\sigma$ are the mean and standard deviation of the rewards within the group, respectively.

However, within the workflow optimization process, scoring an individual workflow involves code execution and potentially unstable API calls. Employing the original GRPO methodology, which requires generating and evaluating workflows on-the-fly during training followed immediately by gradient updates, introduces a critical point of failure. If any single code execution or API call fails or hangs during a training step, the entire training process is forced to halt. Consequently, the stability of each individual code run and API invocation directly impacts the overall stability of the training procedure, rendering this approach infeasible for practical implementation.

% To resolve this fundamental issue, we propose a new Iterative-GRPO. Our key insight is to decouple the inherently unstable data collection phase from the stable policy optimization phase. We refactor the monolithic process into two distinct, independent stages: \textbf{1) offline dataset collection} and \textbf{2) offline policy training}. This separation guarantees that the model training procedure is no longer vulnerable to interruptions caused by failures in code execution or external API services.

% In our experiment, annotating workflow samples (based on code representations) with corresponding scores necessitates a multi-step and complex procedure. Each sampled workflow is written to a temporary Python script and executed. During execution, the code logic sequentially invokes various APIs according to the functions of its operators to produce a final response. This response is subsequently passed through an evaluation function to obtain a score, which serves as the reward signal for the workflow. Since this process involves code execution and external API calls, integrating reward acquisition directly within the training loop is highly impractical. It would not only substantially increase training duration but also introduce significant instability into the training process due to potential failures in code execution or API availability, thereby severely limiting its feasibility for practical implementation.

To circumvent these challenges, we propose a variant of GRPO, termed \textbf{Iterative offline GRPO (ioGRPO)}. This method modifies the forward pass of the original GRPO algorithm by decoupling data collection from policy optimization. Specifically, response sampling and reward acquisition are performed as a separate, offline pre-processing step \textit{before} the commencement of training. During the actual training phase, the optimization process directly reads from a pre-collected dataset containing queries $q$, their corresponding sets of responses $\{g_1, ..., g_n\}$, and associated rewards $\{s_1, ..., s_n\}$ to compute the policy gradient loss. 
Furthermore, starting from a base model, we conduct multiple iterative cycles. After each training iteration, a new model checkpoint is saved. This checkpoint serves a dual purpose: 1) it becomes the starting point for the next training iteration, and 2) it acts as the old policy $\pi_{\text{old}}$ for the subsequent round of data collection. This checkpoint is then used to re-sample a new set of responses and acquire their corresponding rewards, refreshing the training dataset for the next iteration. This iterative process effectively decomposes the monolithic GRPO training procedure into two distinct, alternating phases: \textbf{dataset collection} and \textbf{policy update}. This separation successfully eliminates the adverse effects of code execution and API instability on training robustness while simultaneously achieving a significant reduction in overall training time.

% \subsection{The proposed Approach}
\subsection{Advantage-Masked KL Restriction}
% \paragraph{Advantage-Masked KL Divergence Constraint}

According to recent research efforts, including~\citep{yu2025dapo}, the distribution of long-chain reasoning models can undergo substantial divergence from the initial model during training, making such restriction unnecessary.
However, when the reference model is chosen as the checkpoint from the preceding iteration—which is also the model that generated the offline training data for the current round—the role and effect of the KL restriction warrant further analysis. 
In the objective function of the \textbf{ioGRPO},  we add the term $- \mathbb{D}_{\text{KL}}[\pi_\theta ||\pi_{\text{ref}}] $, which discourages the updated policy from diverging too far from the original reference policy.
the KL penalty term in the latter part of the expression can be formulated as:
\begin{equation}
\mathbb{D}_{\text{KL}}[\pi_\theta ||\pi_{\text{ref}}] = \frac{\pi_{\text{ref}}(y_{i,t} | q, y_{i,<t})}{\pi_\theta(y_{i,t} | q, y_{i,<t})} - \log{\frac{\pi_{\text{ref}}(y_{i,t} | q, y_{i,<t})}{\pi_\theta(y_{i,t} | q, y_{i,<t})}} - 1
% \mathbb{E}{q \sim \mathcal{D}} \left[ \text{KL}\left( \pi{\theta}(\cdot | q) , | , \pi_{\text{ref}}(\cdot | q) \right) \right],
\end{equation}
where $\pi_{\theta}$ denotes the new policy being trained and $\pi_{\text{ref}}$ represents the reference policy. This KL restriction term constrains the deviation of the new policy from the reference policy, ensuring the updated policy does not diverge excessively during iterative optimization.
For the iterative offline GRPO framework, the reference policy $\pi_{\text{ref}}$ can be selected either as the initial base model or as the model checkpoint saved from the previous training iteration.

% [XX] during training the long-CoT reasoning model, the model distribution can
% diverge significantly from the initial model, thus this restriction is not necessary.

However, the original objective function applies the KL restriction uniformly to all responses for a given query. Within a pre-collected dataset, each group contains a mix of both high-quality (advantageous) and low-quality (disadvantageous) responses. Applying the restriction to advantageous responses is desirable, as it prevents the new policy from deviating excessively from the high-performing strategies of the reference model. Conversely, applying the same restriction to disadvantageous responses would force the new policy to remain close to the reference model's poor strategies. This latter effect is counterproductive and misaligned with the core objective of reinforcing the generation of high-advantage outputs. 
% To resolve this conflict, we introduce a \textbf{KL mask} mechanism. This mask enables a selective, response-wise application of the KL restriction based on the estimated advantage value. The restriction is activated only for advantageous responses, ensuring the policy improves while maintaining alignment with the reference model's beneficial behaviors, without being unnecessarily anchored to its suboptimal ones.

% In our experiment, the reference model is defined as the model checkpoint saved from the previous training iteration. Consequently, the GRPO training data collected prior to the current iteration is generated by this specific model. 
Our key modification involves linking the advantage values, computed during the ioGRPO objective estimation, directly to the KL restriction. This integration imbues the KL penalty term with an \emph{advantage-aware selectivity}. Specifically, for a given training sample $\tilde{D}_q$ defined in Eq.(\ref{Equation_1}), the advantage value $A_i$ is computed for each response $g_i(q)$, $i\in [n]$. A positive $A_i$ indicates that the corresponding response should be reinforced, whereas a negative $A_i$ suggests that it should not.
Based on this intuition, we introduce a \textbf{Advantage-Masked KL Restriction (mKL)} $m_i$, defined for each response in the sample as:
\begin{equation}
m_i = \mathbb{I}({A_i > 0}),
\label{Equation_2}
\end{equation}
where $\mathbb{I}$ is the indicator function, $1 \leq i \leq n$. The purpose of this mask is to filter the $n$ sampled responses for a given query $q$, selecting only the $l\left(l<n\right)$ \emph{advantageous responses} ($A_i > 0$) for inclusion in the KL penalty calculation. The KL restriction thus only applies to these advantageous responses, effectively ignoring the contributions from the disadvantageous ones. This mechanism ensures that the KL penalty term constrains the new policy $\pi_{\theta}$ towards the \emph{advantageous segments} of the reference policy $\pi_{\text{ref}}$, rather than constraining it against the entirety of $\pi_{\text{ref}}$'s output distribution.
The modified GRPO objective function, incorporating the proposed \textbf{mKL}, is therefore given by:
\begin{align}
&\mathcal{L}_{\text{ioGRPO-mKL}}(\theta) \nonumber
\\
=& \frac{1}{n} \sum_{i=1}^{n} \frac{1}{|y_i|} \sum_{t=1}^{|y_i|} \left\{ \min \left[ r_{i,t}(\theta) \hat{A}_{i,t}, \text{clip}(r_{i,t}(\theta), 1 - \epsilon, 1 + \epsilon) \hat{A}_{i,t} \right] - \beta\cdot {\color{blue}m_i}\cdot\mathbb{D}_{\text{KL}}[\pi_\theta ||\pi_{\text{ref}}] \right\},
% \mathbb{E}{(q, y) \sim \mathcal{D}} \left[ \frac{\pi_{\theta}(y | q)}{\pi_{\text{ref}}(y | q)} A(q, y) - \beta \cdot m(q, y) \cdot \text{KL}\left( \pi_{\theta}(\cdot | q) , | , \pi_{\text{ref}}(\cdot | q) \right) \right],
\end{align}

where $m_i$ is the mask value defined in Eq.~(\ref{Equation_2}), and $\beta$ is a scaling hyperparameter for the penalty term.

%% file: MainText/4Exp.tex
\section{Experiments}
\subsection{Experimental Setup}
\paragraph{Datasets}
We center our evaluation on five publicly available datasets spanning diverse domains such as code generation, mathematics, and question answering. In particular, we employ the entire collections of HumanEval~\citep{chen2021evaluating} and MBPP~\citep{austin2021program}. 
% For GSM8K~\citep{cobbe2021training}, we leverage a subset of 1,319 examples from its test partition, consistent with the procedure in Aflow~\citep{zhang2024aflow}. 
To focus on advanced and complex problems within the MATH dataset, we extract level-5 difficulty questions from the following categories: Combinatorics and Probability, Number Theory, Pre-algebra, and Pre-calculus, mirroring the selection process of~\citep{hong2024data}. For DROP~\citep{dua2019drop} and HotpotQA~\citep{yang2018hotpotqa}, we adhere to the sampling protocols established in \citep{hu2024automated},~\citep{shinn2023reflexion},~\citep{zhang2024aflow}, and \citep{wang2025scoreflow}, randomly drawing 1,000 instances from each. These samples are then partitioned into training sets and test sets using a 1:4 ratio.

\paragraph{Baselines}
Our evaluation includes several manually constructed static workflow baselines: direct LLM calls, Chain of Thought~\citep{wei2022chain}, Self-Consistency CoT (ensembling 5 generated responses)~\citep{wang2022self}, MedPrompt (3 responses with 5 votes)~\citep{nori2023can}, MultiPersona Debate~\citep{wang2023unleashing}, and Self-Refine (executed over 2 rounds)~\citep{madaan2023self}.
% We also include comparisons with code-based representational automated workflow optimization techniques: ADAS~\citep{hu2024automated} and Aflow~\citep{zhang2024aflow}. For both, \texttt{GPT-4o-mini-2024-07-18} serves as the underlying optimization model. The number of iteration rounds in Aflow is set to 20, following the configuration used by~\citep{zhang2024aflow}.
We also include comparisons with state-of-the-art automated workflow optimization techniques based on code representations: ADAS~\citep{hu2024automated} , Aflow~\citep{zhang2024aflow} , and ScoreFlow~\citep{wang2025scoreflow}. The first two methods both utilize \texttt{GPT-4o-mini-2024-07-18} as their underlying optimization model. Following the configuration used in~\citep{zhang2024aflow}, the number of iteration rounds for Aflow is set to 20. For ScoreFlow, \texttt{Qwen2.5-7B-Instruct} is employed as the generator and \texttt{GPT-4o-mini-2024-07-18} as the executor, with the iteration round set to 3.

\paragraph{Models}
In our primary setup, \texttt{Qwen2.5-7B-Instruct}~\citep{yang2025qwen3} serves as the foundational generator model, with inference conducted using vLLM~\citep{kwon2023efficient}. The executor is \texttt{GPT-4o-mini-2024-07-18}, accessible via API with a temperature set to 0. For ablation experiments, we substitute the generator with \texttt{Qwen2.5-3B-Instruct} and \texttt{Qwen2.5-14B-Instruct}, while still keeping \texttt{GPT-4o-mini-2024-07-18} as the executor. All experiments are conducted on four H20 GPUs employing LoRA~\citep{hu2022lora} for efficient fine-tuning.

\paragraph{Evaluation Metrics}
The final performance is measured by task solve rates, averaged over 3 independent evaluation runs. To mitigate formatting inconsistencies across outputs, \texttt{GPT-4o-mini-2024-07-18} serves as the judge model for the MATH, DROP, and HotpotQA datasets. During each of the 3 optimization iterations, we generate 16 candidate workflows per problem (i.e. $m = 16$) and compute their scores. Meanwhile, the initial group size is set to 14 (i.e. $n = 14$), and the final group size (after group sharpening) is set to 8 (i.e. $t = 4$).
To limit computational expense, no dedicated judge model is employed at this stage. Evaluation relies on F1 scores for DROP and HotpotQA, and on solve rates—also averaged over three runs—for all other benchmarks.

\subsection{Main Results}

\newcommand{\uparrowtext}[1]{\textcolor{green!30!black}{\raisebox{0.1em}{\fontsize{7pt}{7pt}\selectfont ↑}\raisebox{0.05em}{\fontsize{7pt}{7pt}\selectfont#1}}}
\newcommand{\downarrowtext}[1]{\textcolor{red}{\raisebox{0.1em}{\fontsize{7pt}{7pt}\selectfont ↓}\raisebox{0.05em}{\fontsize{7pt}{7pt}\selectfont#1}}}
\definecolor{sigmaBG}{HTML}{D9FFDD}
\definecolor{baselineBG}{HTML}{EDEDED}
\definecolor{SequentialBG}{HTML}{D3F3FE}
\begin{table*}[t]
\caption{Main experimental results comparing SPOGW with baseline methods across five benchmarks: MATH (math reasoning), HumanEval and MBPP (coding), HotpotQA and DROP (question answering). SPOGW achieves state-of-the-art performance on all tasks, with improvements over the previous best method (ScoreFlow) indicated by up arrows.}
\makebox[\textwidth][c]{%
\centering
\small
\setlength{\tabcolsep}{3.5pt}
\renewcommand{\arraystretch}{0.95}
\begin{tabular}{@{\hspace{0pt}}>{\raggedright\arraybackslash}p{3cm}cccccc@{}}
\toprule
\multirow{2}{*}{\textbf{Methods}} 
    & \multicolumn{1}{c}{\textbf{Math Reasoning}} 
      & \multicolumn{2}{c}{\textbf{Coding}} 
        & \multicolumn{2}{c}{\textbf{Question Answering}} 
            & \multirow{2}{*}{\textbf{AVG}} \\
\cmidrule(lr){2-2} \cmidrule(lr){3-4} \cmidrule(lr){5-6}
    & MATH 
      & HumanEval & MBPP 
        & HotpotQA & DROP 
            & \\
\midrule
IO                                     &52.2 &90.1 &69.5 &73.6 &81.6 &73.4 \\
CoT~\citep{wei2022chain}                &53.4 &91.6 &70.4 &73.4 &83.2 &74.4 \\
CoT SC~\citep{wang2022self}             &53.8 &92.9 &71.3 &74.0 &83.2 &75.0 \\
MedPrompt~\citep{nori2023can}           &53.7 &92.1 &69.2 &74.4 &83.0 &74.5 \\
MultiPersona~\citep{wang2023unleashing} &51.9 &92.9 &70.4 &73.1 &81.3 &73.9 \\
Self Refine~\citep{madaan2023self}      &50.0 &91.1 &70.0 &73.6 &82.5 &73.4 \\
ADAS~\citep{hu2024automated}            &51.7 &88.8 &68.7 &78.5 &81.3 &73.8 \\
Aflow~\citep{zhang2024aflow}            &55.8 &92.9 &82.9 &77.9 &83.5 &78.6 \\
\rowcolor{SequentialBG}
ScoreFlow~\citep{wang2025scoreflow}     &60.0 &95.1 &83.2 &84.1 &84.3 &81.3 \\
\rowcolor{sigmaBG}
\textbf{SPOGW(Ours)}                    &\textbf{62.3}\uparrowtext{2.3} 
                                       &\textbf{96.2}\uparrowtext{1.1} &\textbf{84.1}\uparrowtext{0.9} &\textbf{85.0}\uparrowtext{0.9} &\textbf{85.3}\uparrowtext{1.0} &\textbf{82.6}\uparrowtext{1.3} \\
\bottomrule
\end{tabular}
}
\label{tab:main-results}
\end{table*}

\begin{table*}[t]
\caption{Performance comparison of different generator models with and without SPOGW optimization on HumanEval and HotpotQA. SPOGW enables smaller models to achieve performance competitive with larger baseline models.}
\makebox[\textwidth][c]{%
\centering
\small
\setlength{\tabcolsep}{4pt}
\renewcommand{\arraystretch}{0.95}
\begin{tabular}{@{\hspace{0pt}}>{\raggedright\arraybackslash}p{4.5cm}cc@{}}
\toprule
\textbf{Generator Model} & \textbf{HumanEval} & \textbf{HotpotQA} \\
\midrule
\rowcolor{SequentialBG}
Qwen2.5-3B-Instruct           &91.9  &84.1 \\
\rowcolor{SequentialBG}
Qwen2.5-7B-Instruct           &93.4  &84.2 \\
Qwen2.5-14B-Instruct          &94.4  &84.7 \\
\midrule\noalign{\vskip -\belowrulesep}
\rowcolor{sigmaBG}
\textbf{Qwen2.5-3B-Instruct-SPOGW} &\textbf{94.1}\uparrowtext{2.2}  &\textbf{84.3}\uparrowtext{0.2} \\
\rowcolor{sigmaBG}
\textbf{Qwen2.5-7B-Instruct-SPOGW} &\textbf{96.2}\uparrowtext{2.8}  &\textbf{85.0}\uparrowtext{0.8} \\
\bottomrule
\end{tabular}
}
\label{tab:generator models}
\end{table*}

We present the main experimental results comparing SPOGW against a comprehensive set of baseline methods across five benchmark datasets spanning mathematical reasoning, coding, and question answering domains. As shown in Table~\ref{tab:main-results}, SPOGW consistently achieves state-of-the-art performance, outperforming all baseline methods on every benchmark.

Specifically, In \textbf{mathematical reasoning (MATH)}, SPOGW attains a solve rate of \textbf{62.3$\%$}, surpassing the previous best method, ScoreFlow, by \textbf{2.3} percentage points. This demonstrates SPOGW's effectiveness in handling complex, multi-step reasoning tasks requiring structured problem-solving workflows.
For \textbf{code generation tasks,} SPOGW achieves \textbf{96.2$\%$} on HumanEval and \textbf{84.1$\%$} on MBPP, exceeding ScoreFlow by \textbf{1.1} and \textbf{0.9} percentage points, respectively. This improvement highlights SPOGW's capability in generating functionally correct code through optimized workflow structures.
In \textbf{question answering}, SPOGW obtains \textbf{85.0$\%$} on HotpotQA and \textbf{85.3$\%$} on DROP, representing gains of \textbf{0.9} and \textbf{1.0} percentage points over ScoreFlow. These results indicate that SPOGW effectively handles multi-hop reasoning and discrete reasoning over textual content.

Across all benchmarks, SPOGW achieves an average performance of \textbf{82.6$\%$}, a \textbf{1.3} percentage point improvement over the previous state-of-the-art. Notably, SPOGW not only outperforms automated workflow optimization methods (ADAS, Aflow, ScoreFlow) but also exceeds carefully designed manual workflows such as MedPrompt, MultiPersona, and Self-Refine.
The consistent superiority of SPOGW across diverse domains underscores the effectiveness of our group-wise preference optimization approach. The improvements are particularly significant in mathematical reasoning, where the structured nature of workflows plays a crucial role in solving complex problems. These results validate SPOGW as a robust and general-purpose framework for automated workflow generation and optimization.

\subsection{Ablation Studies}

\textbf{Analysis of the generator model}
As shown in Table~\ref{tab:generator models}, we investigate the impact of the generator model by comparing Qwen2.5 models of varying sizes on HumanEval and HotpotQA. The results show that SPOGW optimization not only improves the performance of the original model but even effectively compensates for the limitations of the model scale: while baseline performance improves with increasing model size (e.g., HumanEval scores rising from 91.9 to 94.4), SPOGW-enhanced smaller models achieve performance comparable to or even surpassing larger baseline models. Specifically, Qwen2.5-3B-Instruct-SPOGW attains 94.1 on HumanEval, closely approaching the baseline 14B model's 94.4, while Qwen2.5-7B-Instruct-SPOGW reaches 96.2, exceeding all baseline models including the 14B variant. This indicates that SPOGW's group-wise optimization effectively amplifies the capabilities of smaller models, reducing dependency on model scale while maintaining strong performance across reasoning tasks.

\begin{table*}[t]
\caption{Ablation study on KL restriction configurations. Results show that combining the iterative checkpoint with selective KL mask yields the highest performance.
% validating the design of SPGW's advantage-aware restriction mechanism.
}
\makebox[\textwidth][c]{%
\centering
\small
\setlength{\tabcolsep}{4pt}
\renewcommand{\arraystretch}{0.95}
\begin{tabular}{@{\hspace{0pt}}>{\raggedright\arraybackslash}p{3cm}p{4.5cm}cc@{}}
\toprule
\textbf{Objective Function} & \textbf{Reference Model} & \textbf{Enable KL Mask} & \textbf{HumanEval} \\
\midrule
w/o KL term  &None                & \ding{55}  &94.9 \\
\midrule\noalign{\vskip -\belowrulesep}
w/ KL term   &Initial base model  & \ding{55}  &94.4 \\
\rowcolor{SequentialBG}
w/ KL term   &Previous iteration's checkpoint  & \ding{55}   &95.4 \\
\rowcolor{sigmaBG}
\textbf{w/ KL term}     &\textbf{Previous iteration's checkpoint}  &\ding{51}    &\textbf{96.2}\uparrowtext{0.8} \\
\bottomrule
\end{tabular}
}
\label{tab:KL restriction}
\end{table*}

\begin{table*}[t]
\caption{Impact of data processing methods on HumanEval performance. Progressive refinement from random sampling to screening and sharpening demonstrates the importance of data quality curation for effective policy optimization, with the combined approach yielding the optimal result.}
\makebox[\textwidth][c]{%
\centering
\small
\setlength{\tabcolsep}{4pt}
\renewcommand{\arraystretch}{0.95}
\begin{tabular}{@{\hspace{0pt}}>{\raggedright\arraybackslash}p{2cm}p{4.5cm}cc@{}}
\toprule
\textbf{Dataset} & \textbf{Data Processing Methods} & \textbf{Filtering First} & \textbf{HumanEval} \\
\midrule
$\mathcal{D}_\text{RS}$ &Random Sampling           &\ding{55}  &93.4 \\
$\mathcal{D}_\text{RSF}$ &Random Sampling           &\ding{51}  &94.9 \\
$\mathcal{D}_\text{S}$ &Only Screening                 &\ding{51}  &95.7 \\
\rowcolor{sigmaBG}
$\mathcal{D}_\text{SS}$ &\textbf{Screening and Sharpening} &\ding{51}  &\textbf{96.2} \\
\bottomrule
\end{tabular}
}
\label{tab:data processing methods}
\end{table*}

\begin{figure}[t]
\centering
\subfigure[Variance distribution]{\includegraphics[height=3.5cm,width=6cm]{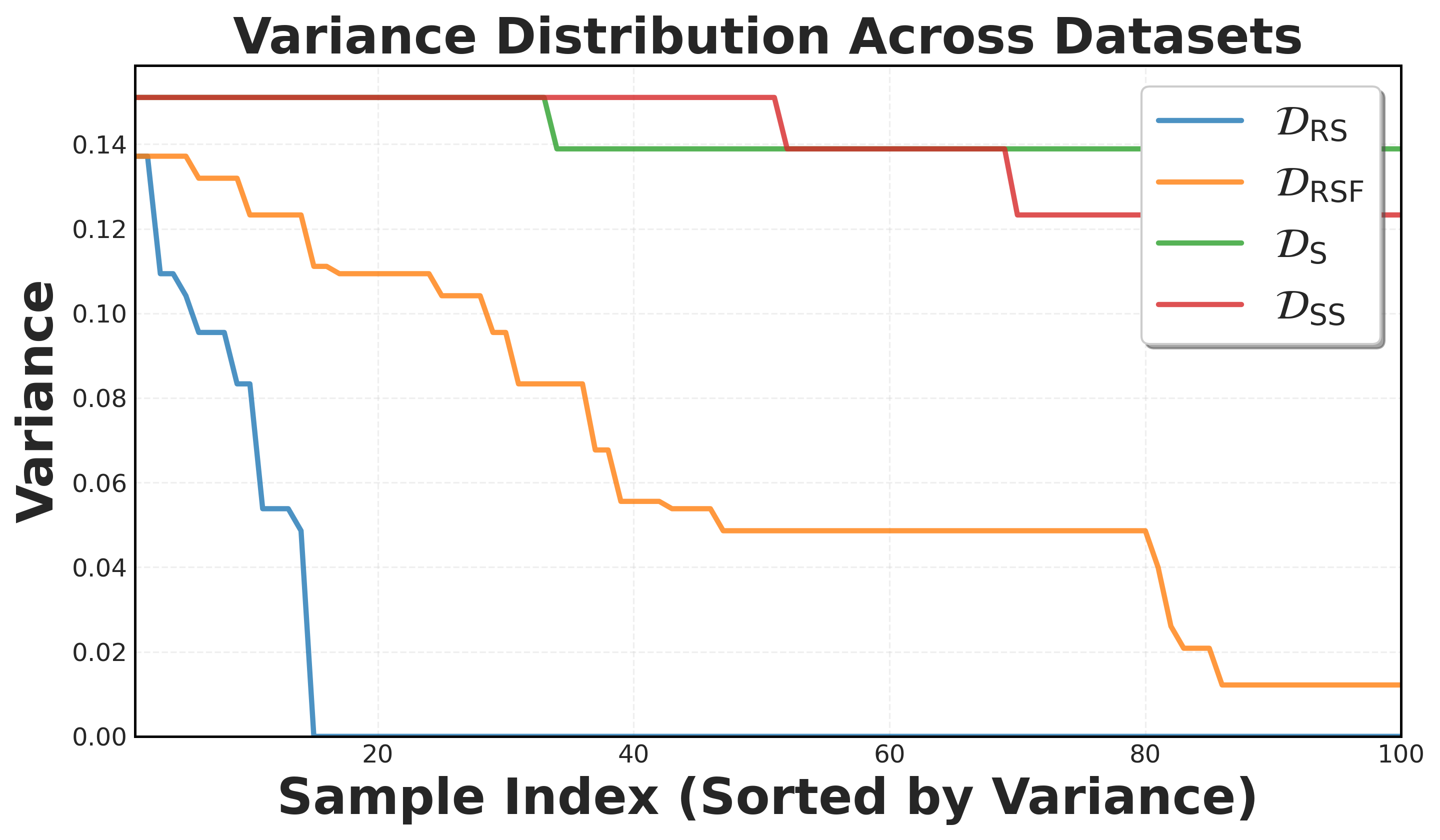}}
\hspace{0.4cm}
\subfigure[Median interval length distribution.]{\includegraphics[height=3.5cm,width=6cm]{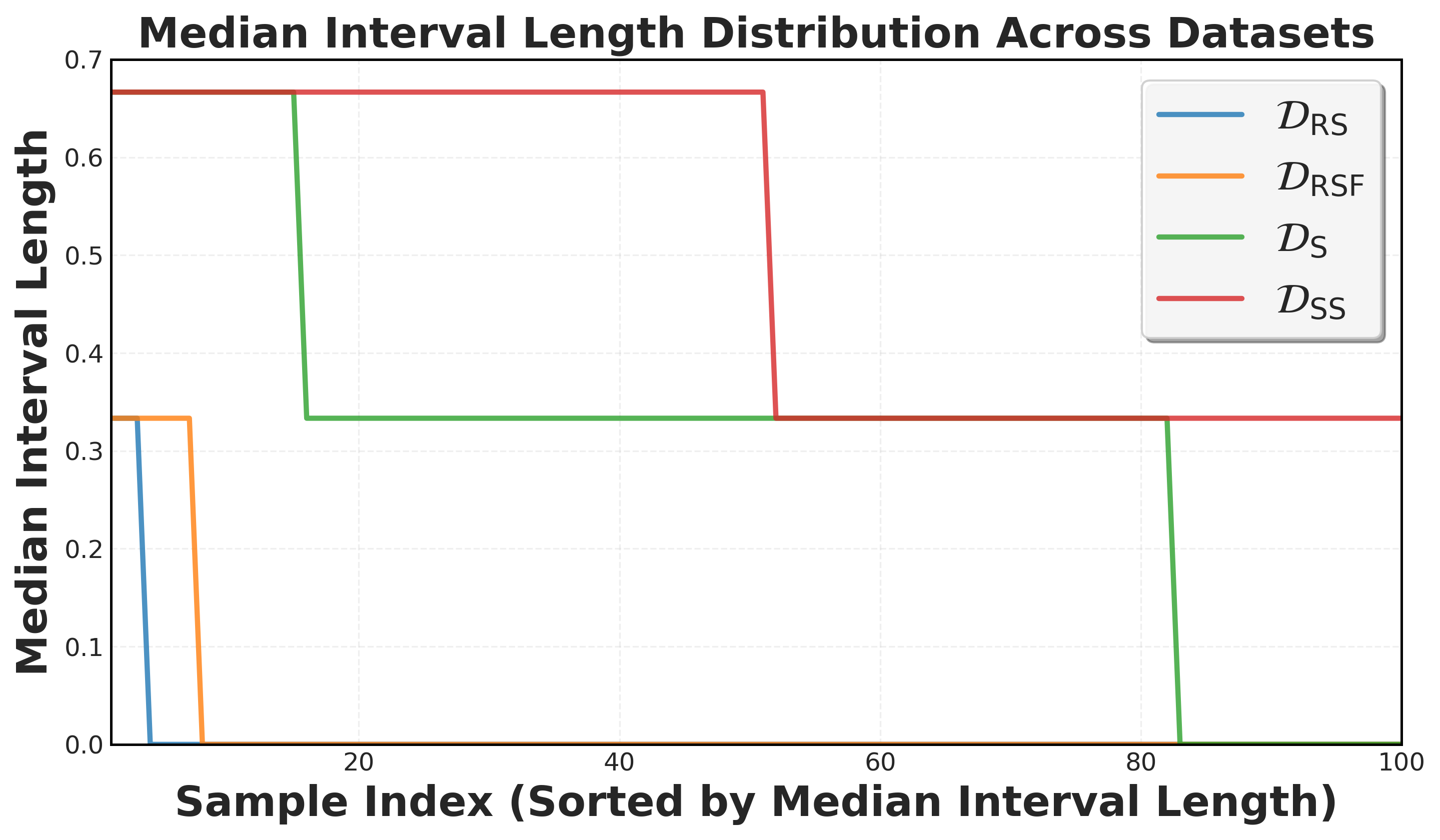}}
\caption{
% Analysis of dataset characteristics for different processing methods. Dataset 4 (screening and sharpening) exhibits superior variance properties and quality separation, explaining its enhanced performance in policy optimization. Here, the training group size is fixed at 8. The \textbf{Median Interval Length (MIL)} is defined as the difference between the 4th and 5th highest scores in a sample. A higher MIL indicates that the sample is more polarized, with a clearer distinction among high-quality and low-quality responses.
Analysis of dataset characteristics under different processing methods shows that $\mathcal{D}_\text{SS}$ achieves superior variance and clearer quality separation.
% leading to better policy optimization. 
The training group size is fixed at 8. Median Interval Length (MIL) is the gap between the 4th and 5th highest scores.
% a larger MIL indicates greater polarization and clearer distinction between high- and low-quality responses.
}
\label{fig:data_processing_2}
\end{figure}
\begin{figure}[t]
\centering
\subfigure[]{\includegraphics[height=2.7cm,width=4cm]{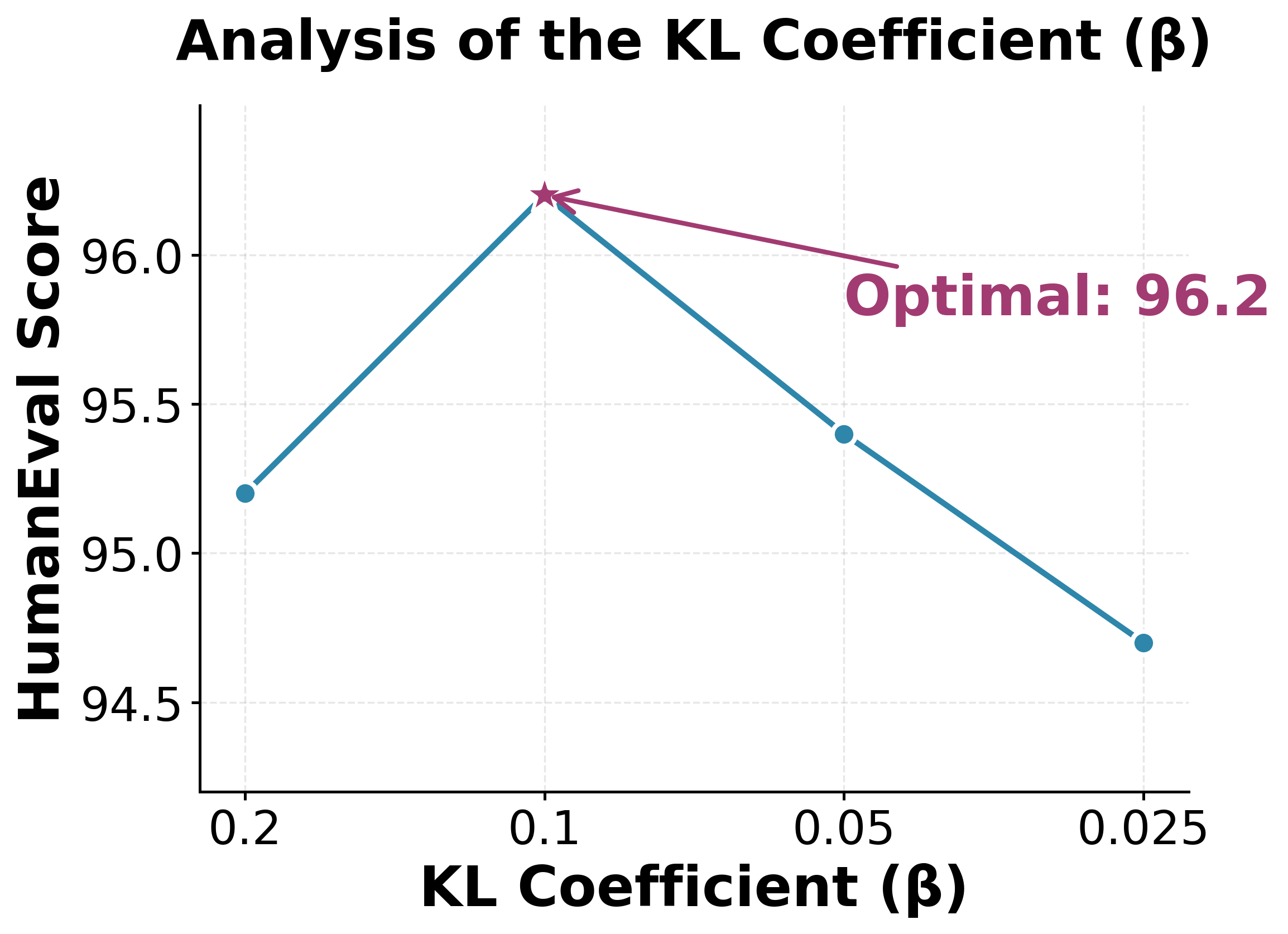}\label{fig:kl_coefficient}}
\hspace{0.1cm}
\subfigure[]{\includegraphics[height=2.7cm,width=4cm]{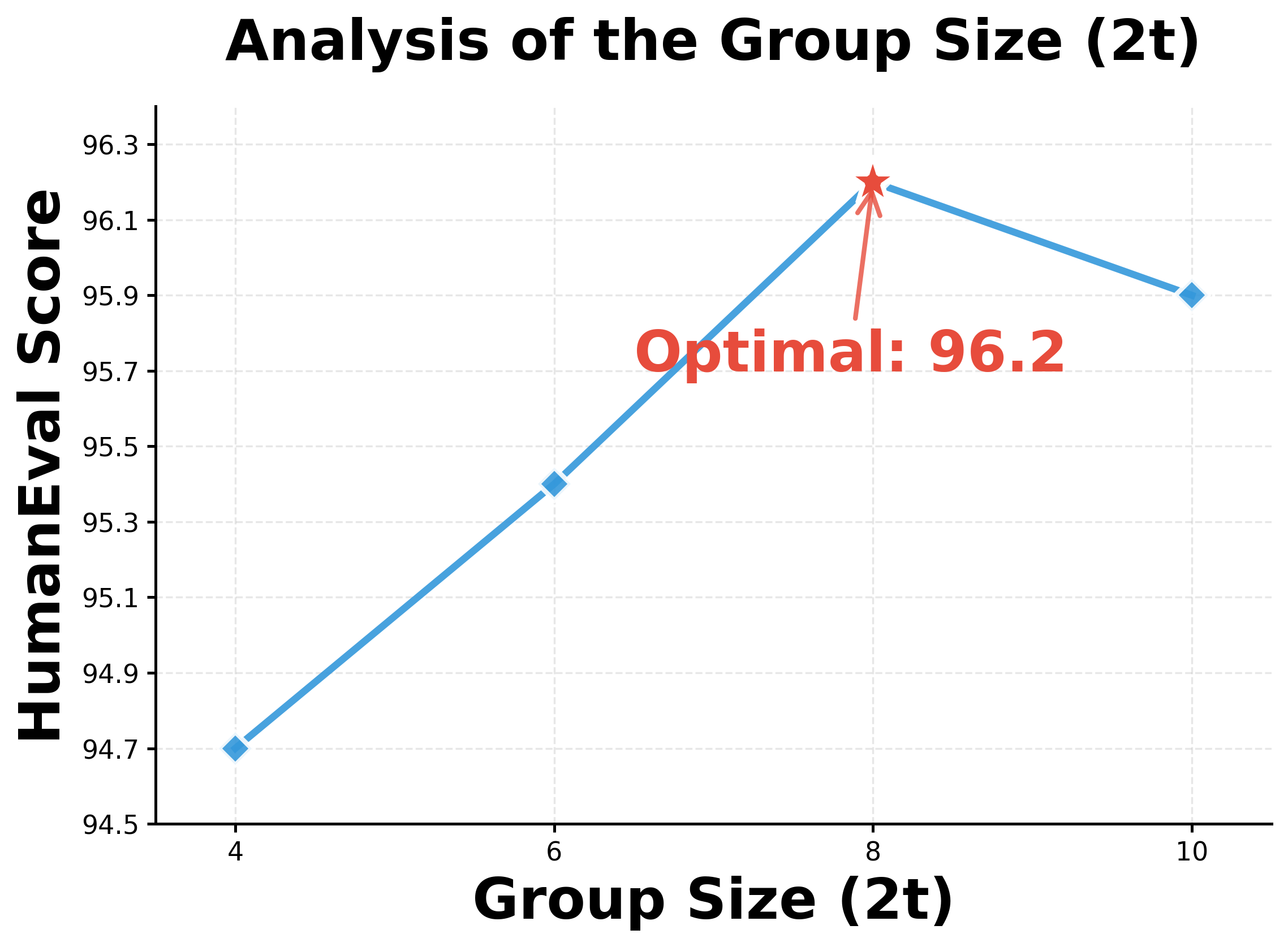}\label{fig:group_size}}
\hspace{0.1cm}
\subfigure[]{\includegraphics[height=2.7cm,width=4cm]{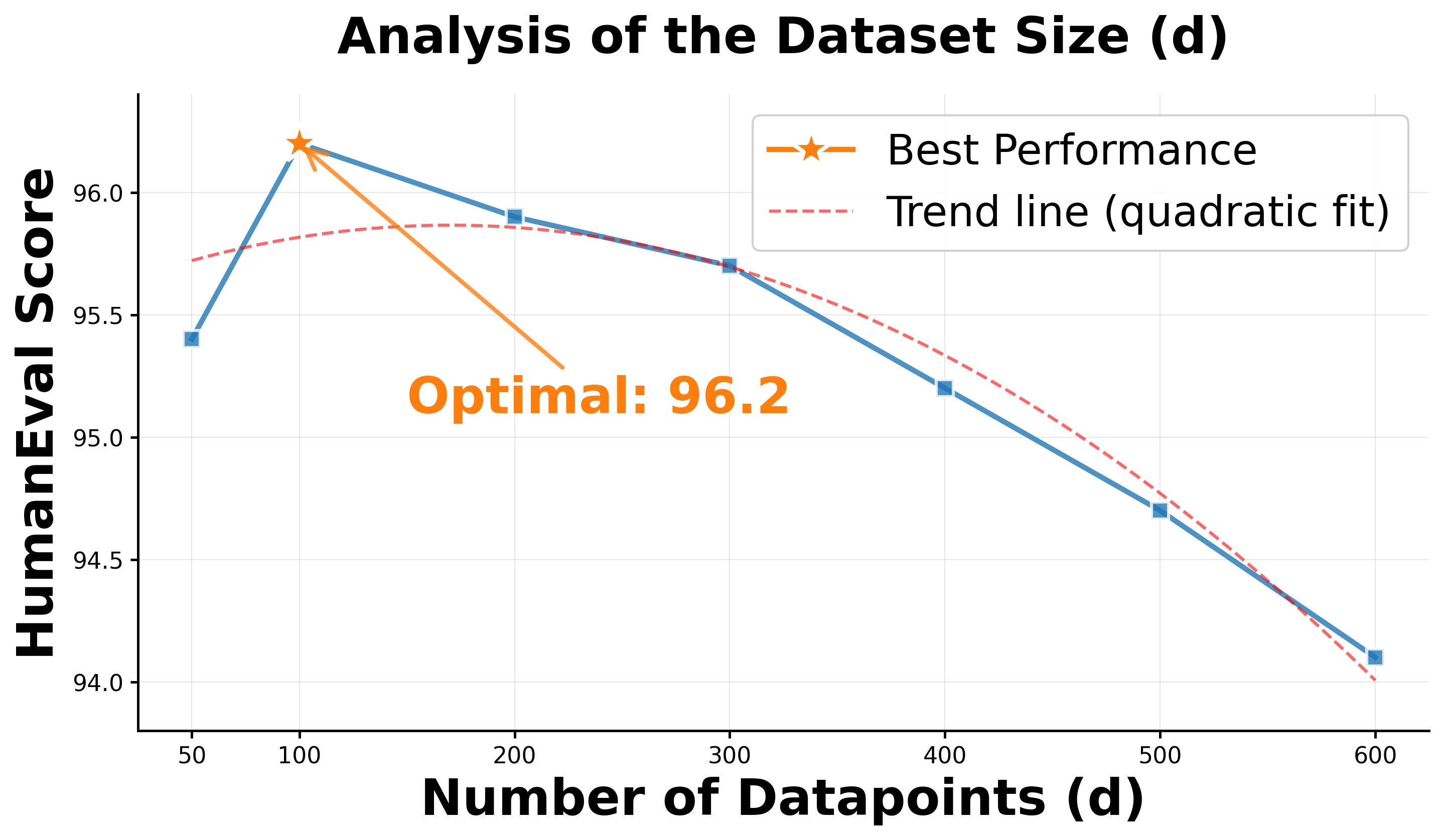}\label{fig:dataset_size}}
\vskip -0.1in
\caption{Analysis of the KL coefficient $\beta$, group size $2t$ and dataset size $d$ on HumanEval. 
% The experiment investigates the effect of varying each of these parameters individually and identifies their optimal values.
}
\label{fig:data_processing}
\end{figure}

\textbf{Analysis of the KL restriction}
As shown in Table~\ref{tab:KL restriction}, we ablate the impact of the KL restriction and the proposed advantage-masked mechanism on HumanEval performance. Removing the KL term entirely yields a score of 94.9, while applying KL regularization with the initial base model as reference degrades performance to 94.4, indicating that rigid constraint towards an outdated policy can hinder optimization. Switching the reference model to the previous iteration’s checkpoint improves results to 95.4, demonstrating the benefit of iterative policy alignment. Finally, enabling the KL mask—which selectively applies KL penalty only to advantageous responses—further boosts performance to 96.2, underscoring that targeted restriction towards high-quality behaviors is crucial for stable and effective policy improvement.

\textbf{Analysis of the data processing method}
As demonstrated in Table~\ref{tab:data processing methods} and Figure~\ref{fig:data_processing_2}, the progressive refinement of data processing methods significantly enhances model performance on HumanEval, with random sampling achieving 93.4, filtering improving to 94.9, screening alone reaching 95.7, and the combined screening and sharpening approach yielding the optimal 96.2. The variance and median interval length distributions reveal that Dataset 4 exhibits both higher variance and clearer separation among reward scores, confirming that our curated data processing pipeline effectively amplifies intra-group diversity and reward distinction, thereby providing stronger and clearer learning signals for advantage estimation and policy optimization.

\textbf{Analysis of the KL coefficient $\beta$, the group size, and the dataset size}
Our ablation study on the KL coefficient ($\beta$), group size (2t), and dataset size (d) highlights the importance of balanced hyperparameters for stable and efficient policy optimization. As shown in Figures \ref{fig:kl_coefficient}–\ref{fig:dataset_size}, excessive KL regularization ($\beta$ = 0.2) restricts exploration while insufficient regularization ($\beta$ = 0.025) destabilizes learning, with the optimal $\beta$ = 0.1 achieving the highest score (96.2). Similarly, SPOGW reaches peak performance at a group size of 2t = 8, where smaller or larger groups either reduce contrast for advantage estimation or introduce noise. Performance also follows an inverted U-shaped trend with respect to dataset size, peaking at d = 100 and declining for both smaller and larger datasets. These results collectively underscore the critical role of properly tuned regularization strength, group size, and data scale in maintaining sharp reward distinctions and preventing overfitting, thereby maximizing policy performance.

%% file: MainText/6Con.tex
\section{Conclusion}
We present SPOGW, a score-based preference optimization method for automated agentic workflow generation that overcomes the limits of discrete optimization and pairwise comparisons via group-wise optimization in continuous space. SPOGW introduces three innovations: 1) variance-based group-wise data construction, 2) an iterative offline GRPO framework decoupling data collection from policy updates for stability, and 3) an advantage-masked KL restriction guiding policy divergence toward high-quality behaviors. Experiments on reasoning, coding, and QA benchmarks show SPOGW surpasses state-of-the-art methods, while ablations confirm each component’s contribution and highlight optimal hyperparameter settings. SPOGW offers a scalable, effective framework that reduces manual design while maintaining strong performance across domains.